\documentclass[]{article}

\usepackage{graphicx}
\usepackage{subfig}
\usepackage{amsmath}
\usepackage{amssymb}
\usepackage{booktabs} 

\usepackage{natbib}

\title{Weekly maintenance scheduling using exact and genetic methods}
\author{Andrew W. Palmer, Robin Vujanic, Andrew J. Hill, Steven J. Scheding}

\begin{document}

\maketitle

\begin{abstract}
The weekly maintenance schedule specifies when maintenance activities should be performed on the equipment, taking into account the availability of workers and maintenance bays, and other operational constraints. The current approach to generating this schedule is labour intensive and requires coordination between the maintenance schedulers and operations staff to minimise its impact on the operation of the mine. This paper presents methods for automatically generating this schedule from the list of maintenance tasks to be performed, the availability roster of the maintenance staff, and time windows in which each piece of equipment is available for maintenance. Both Mixed-Integer Linear Programming (MILP) and genetic algorithms are evaluated, with the genetic algorithm shown to significantly outperform the MILP. Two fitness functions for the genetic algorithm are also examined, with a linear fitness function outperforming an inverse fitness function by up to 5\% for the same calculation time. The genetic algorithm approach is computationally fast, allowing the schedule to be rapidly recalculated in response to unexpected delays and breakdowns. 
\end{abstract}

\section{Introduction}

Maintenance activities are a significant cost in the mining sector, making up between 30\% and 60\% of the total operating cost of a mine \citep{Singleton1998,Lewis2001,Dhillon2008}. In addition to direct costs, poorly planned maintenance can have a large impact on mine productivity---equipment spending longer in maintenance than needed effectively reduces the total tonnes of ore that a mine can produce. As a result, generating maintenance schedules that minimise the impact on the operation of the mine is of great importance to mine operators. Existing work on planning maintenance in the mining industry has looked at modelling the reliability of the equipment \citep{Summit2015}, or focused on long timescales, allocating equipment to tasks with the objective to minimise the expected maintenance costs over the lifetime of the mine \citep{Topal2010}. This paper examines the problem of generating the weekly maintenance schedule for the multitude of activities to be performed on each piece of equipment, with the objective of minimising the impact of the maintenance on the mine operations. The proposed approach is aware of constraints such as the number of maintenance bays and the availability roster of the maintainers, and is demonstrated on real and simulated datasets that reflect practical problem sizes. 

The current approach to producing weekly maintenance schedules is largely manual, and requires the dedicated schedulers to coordinate with the operations planners to develop a schedule that satisfies operational constraints. While the intention is that the schedule should be generated once for the week, they are frequently revised to incorporate unexpected breakdown of equipment and delays \citep{Tomlingson2007}. The aim of this work is to automatically and quickly generate the schedule given the list of maintenance tasks that are to be performed, the availability roster of the maintenance staff, and time windows for each piece of equipment in which the equipment can be taken down for maintenance with minimal impact on the mine operations. 

To the best of the authors' knowledge, the weekly maintenance scheduling problem has not been studied in the literature. Some related maintenance scheduling problems are examined by \cite{Gopalakrishnan2001, Deris1999, BenAli2011, Jiu2013, Jin2009, Aissani2009, Pandey2011, Najid2011}. Many of these authors used meta-heuristics such as tabu-search \citep{Gopalakrishnan2001} and Genetic Algorithms (GAs) \citep{Deris1999,BenAli2011,Jiu2013} to generate a schedule. These meta-heuristic approaches were shown to produce near-optimal solutions in reasonable calculation times. An option-based cost model was used by \cite{Jin2009}, while a novel reinforcement learning approach was utilised by \cite{Aissani2009}. Methods for jointly optimising maintenance and production were developed by \cite{Pandey2011, Najid2011}. \cite{Najid2011} formulated this as a Mixed-Integer Linear Program (MILP) and solved it using commercial optimisation software. However, the authors pointed out that only small problem instances were solvable, and stated that they intend to investigate meta-heuristics for solving larger instances. 

The major difference between the above problems and producing a weekly mine maintenance schedule is in the size of the problem. Typical weekly maintenance schedules can contain over 100 pieces of equipment, with up to 50 individual activities per piece of equipment. Durations for each activity can range from half an hour to several days, and approximately 25 different types of workers, such as fitters, electricians, and boilermakers, are required. The availability of these workers varies both day to day and between the day and night shifts. One of the characteristics of mining in general is that it is very dynamic, with unexpected events such as equipment failures occurring frequently. Thus, one of the requirements of the scheduling system is that it should be computationally fast to allow rapid replanning in response to these unexpected events. 

Two approaches to this problem are proposed in this paper---a MILP formulation that is solved using Gurobi \citep{Gurobi}, and a GA. While the MILP approach generates optimal solutions, it will be shown to be computationally infeasible for realistic problem sizes, motivating the use of the GA. The specific contributions of this paper are:

\begin{itemize}
\item A MILP formulation of the problem
\item A GA approach that utilises a greedy heuristic to ensure that feasible schedules are generated from the chromosomes
\item A comprehensive evaluation of the MILP method and the GA using two different fitness functions
\end{itemize}

The remainder of this paper is structured as follows: Section \ref{s:problem} presents a MILP formulation of the problem, and Section \ref{s:genetic} develops the GA. A comparison of the methods is presented in Section \ref{s:comp_study}, and concluding remarks are provided in Section \ref{s:conc}.

\section{MILP Formulation} \label{s:problem}

\begin{figure}
	\centering
	
	\includegraphics[width=\textwidth]{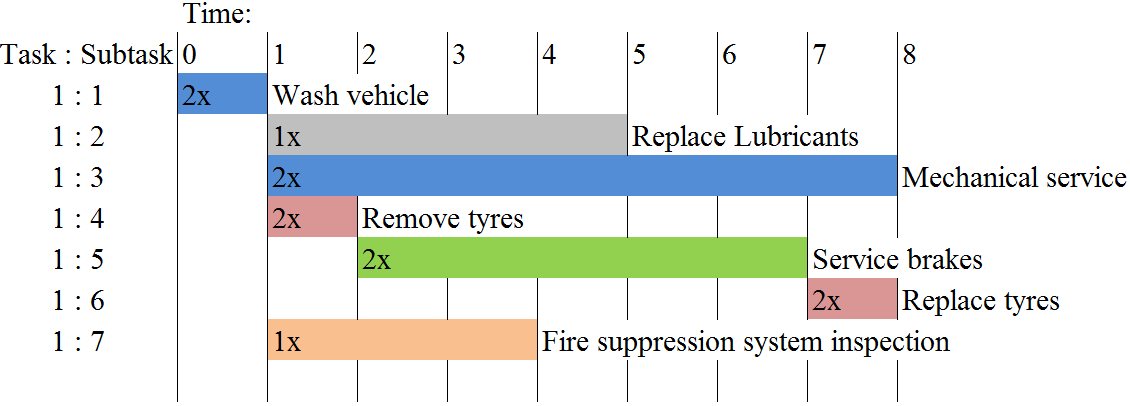}
	
	\caption{An example task with its subtasks scheduled as early as possible taking into account the precedence constraints. Examples of the precedence constraints seen in this example are the vehicle must be washed before further tasks are performed, and the tyres must be removed and replaced before and after the brake service. The colour of the subtask specifies the type of personnel required (5 in the example, which could include electrician, fitter, boilermaker, etc.), and the number of people required is specified by the number in each subtask.}
	\label{f:sample}
\end{figure}

The weekly maintenance problem is to schedule a given set of tasks (equipment) $i \in I$ with associated subtasks (work orders) $j \in J_{i}$. Each task has a ready time that specifies the earliest a task can be started, and a deadline that specifies when it should be completed by. Some tasks require a maintenance bay, and subtasks have a specified duration and worker requirement. Subtasks can have a precedence requirement where they require other subtasks of the task to be performed before it can be performed. An example task with 7 subtasks is shown in Figure \ref{f:sample}. The worker requirements of the subtasks in the example are denoted by the colours and specified numbers in the subtasks, and the subtasks have been scheduled to respect the precedence of the subtasks. The schedule is split into discrete, equally-sized time periods. The objective of the problem is to schedule the subtasks to minimise the sum of the makespan and lateness of the tasks. Makespan is simply the length of a task, and lateness is incurred if the task is completed after the deadline. The rest of this section formally presents the MILP model. 

\subsection{Indices and Sets}

\begin{tabular}{p{2.5cm} p{9cm}}
$t \in T$ & time periods, where $ T = \{0,\dots,t_{max}\}$ \\
$i \in I$ & tasks \\
$j \in J$ & subtasks \\
$p \in P$ & worker types \\
\end{tabular}

The set of tasks that require a maintenance bay is $M \subseteq I$, the set of subtasks that compose task $i$ is $J_{i} \subseteq J$, and the set of subtasks that must be performed before $j$ is $K_{j} \subseteq J$. Subtasks can only belong to one task, so $J_{i} \cap J_{k} = \emptyset \; \forall i \in I, k \in I\backslash\{i\}$.

\subsection{Parameters}

\begin{tabular}{p{2.5cm} p{9cm}}
$d_{j}$ & the duration of subtask $j \in J$ in time periods \\
$r_{j,p,s,t}$ & the number of workers of type $p \in P$ required for subtask $j \in J$ in timestep $t \in T$ if $j$ started in timestep $s \in \{\max(0,t-d_{j},\dots,t)\}$\\
$a_{p,t}$ & the number of workers of type $p \in P$ available in time period $t \in T$ \\
$m$ & the number of maintenance bays \\
$b_{i}$ & ready time for task $i \in I$\\
$c_{i}$ & deadline for task $i \in I$\\
$f_{i}$ & the weighting of the makespan of task $i \in I$ \\
$g_{i}$ & the weighting of the lateness of task $i \in I$ \\
\end{tabular}

\subsection{Optimisation Variables} \label{s:variables}

\begin{tabular}{p{2.5cm} p{9cm}}
$x_{j,t}^{\textrm{start}} \in \{0,1\}$ & 1 iff subtask $j \in J$ starts in time period $t \in T$\\
$x_{i,t} \in \{0,1\}$ & 1 iff task $i \in I$ is being performed in time period $t \in T$\\
$x_{i,t}^{\textrm{start}} \in \{0,1\}$ & 1 iff task $i \in I$ starts in time period $t \in T$\\
$x_{i,t}^{\textrm{finish}} \in \{0,1\}$ & 1 iff task $i \in I$ finishes in time period $t \in T$\\
$y_{i}^{\textrm{makespan}} \in \mathbb{R}$ &  makespan of task $i \in I$\\
$y_{i}^{\textrm{lateness}} \in \mathbb{R}$ & lateness of task $i \in I$\\

\end{tabular}

\subsection{Objective function}

The objective function is the sum of the weighted makespan and lateness of each task:

\begin{equation}\label{eq:objective}
J = \sum\limits_{i \in I} f_{i} y_{i}^{\textrm{makespan}} + g_{i} y_{i}^{\textrm{lateness}}
\end{equation}

\subsection{Constraints} \label{s:constraints}

\subsubsection{Subtasks}

The first set of constraints specify when a subtask can start. Firstly, (\ref{eq:sum_to_1}) enforces that each subtask has to start exactly once, and (\ref{eq:sub_duration}) specifies that it has to be completed before the end of the schedule. Constraint (\ref{eq:precedence}) encodes the precedence constraints as specified in $K_{j}$. 

\begin{equation} \label{eq:sum_to_1}
\sum\limits_{t}x_{j,t}^{\textrm{start}} = 1 \qquad \forall j \in J
\end{equation}

\begin{equation} \label{eq:sub_duration}
\sum\limits_{t}(tx_{j,t}^{\textrm{start}}) + d_{j} \le t_{\max} + 1 \qquad \forall j \in J
\end{equation}

\begin{equation} \label{eq:precedence}
\sum\limits_{t}(tx_{j,t}^{\textrm{start}}) \ge \sum\limits_{t}(tx_{k,t}^{\textrm{start}}) + d_{k} \qquad \forall j \in J, k \in K_{j}
\end{equation}

\subsubsection{Tasks}

The next set of constraints deal with the tasks. Constraint (\ref{eq:sum_to_1_task}), similar to (\ref{eq:sum_to_1}), specifies that a task can only start once. The period that the task starts in is defined in (\ref{eq:task_start}) as the earliest starting period of its subtasks. The makespan of the task is calculated in (\ref{eq:task_makespan}) by specifying that the task must finish no earlier than any of its subtasks. Constraint (\ref{eq:task_finish}) enforces that the task can only finish once, and (\ref{eq:task_finish2}) specifies that the period in which the task is finished in is no earlier than its start period plus its makespan. The lateness of the task is calculated in (\ref{eq:task_lateness}). Constraint (\ref{eq:task}) calculates whether a task is being performed in a time period or not based on whether it was being performed in the previous time period, and the time period in which the task starts and finishes. For (\ref{eq:task}) to work when $t=0$, a dummy variable is added for each task in the time period $t=-1$ and assigned the value of 0 in (\ref{eq:initial_task}).

\begin{equation} \label{eq:sum_to_1_task}
\sum\limits_{t}x_{i,t}^{\textrm{start}} = 1 \qquad \forall i \in I
\end{equation}

\begin{equation} \label{eq:task_start}
\sum\limits_{t}(tx_{i,t}^{\textrm{start}}) \le \sum\limits_{t}(tx_{j,t}^{\textrm{start}}) \qquad \forall i \in I, j \in J_{i}
\end{equation}

\begin{equation} \label{eq:task_makespan}
\sum\limits_{t}(tx_{i,t}^{\textrm{start}}) + y_{i}^{\textrm{makespan}} \ge \sum\limits_{t}(tx_{j,t}^{\textrm{start}}) + d_{j} \qquad \forall i \in I, j \in J_{i}
\end{equation}

\begin{equation} \label{eq:task_finish}
\sum\limits_{t}x_{i,t}^{\textrm{finish}} = 1 \qquad \forall i \in I
\end{equation}

\begin{equation} \label{eq:task_finish2}
\sum\limits_{t}(tx_{i,t}^{\textrm{finish}}) \ge \sum\limits_{t}(tx_{i,t}^{\textrm{start}}) + y_{i}^{\textrm{makespan}} \qquad \forall i \in I
\end{equation}

\begin{equation} \label{eq:task_lateness}
y_{i}^{\textrm{lateness}} \ge \sum\limits_{t}(tx_{i,t}^{\textrm{finish}}) - c_{i} \qquad \forall i \in I
\end{equation}

\begin{equation} \label{eq:task}
x_{i,t} = x_{i,t-1} + x_{i,t}^{\textrm{start}} - x_{i,t}^{\textrm{finish}} \qquad \forall i \in I,t \in T
\end{equation}

\begin{equation} \label{eq:initial_task}
x_{i,-1} = 0 \qquad \forall i \in I
\end{equation}

\subsubsection{Resources}

There are two resource constraints, namely a limited number of maintenance bays, and a limited number of workers. Constraint (\ref{eq:maintenance_bays}) enforces a limit on the number of maintenance bays being used in each time period. Note that only a subset of the tasks require a maintenance bay. Constraint (\ref{eq:person_constraint}) encodes the limit on the number of each type of worker in each time period. 

\begin{equation} \label{eq:maintenance_bays}
\sum\limits_{i \in M} x_{i,t} \le m \qquad \forall t \in T
\end{equation}

\begin{equation} \label{eq:person_constraint}
\sum\limits_{j \in J} \sum\limits_{s \in \{\max(0,t-d_{j}),\dots,t \}} x_{j,s}^{\textrm{start}} r_{j,p,s,t} \le a_{p,t} \qquad \forall p,t
\end{equation}

\subsection{Optimisation Problem}

The maintenance scheduling problem is therefore expressed as the optimisation model $P$:

\begin{equation*}
P = 
\begin{cases}
\text{minimise} & (\ref{eq:objective})\\
\text{subject to} & (\ref{eq:sum_to_1})-(\ref{eq:person_constraint})
\end{cases}
\end{equation*}

Two solution methods to this optimisation problem are proposed---a branch and bound approach using commercial software package Gurobi \citet{Gurobi}, and a GA that is developed in the next section. As will be shown later in the results, this optimisation problem is hard for Gurobi, motivating the development of the genetic algorithm. 

\section{Genetic Algorithm} \label{s:genetic}

\begin{figure}
	\centering
	
	\includegraphics[height=0.7\textheight]{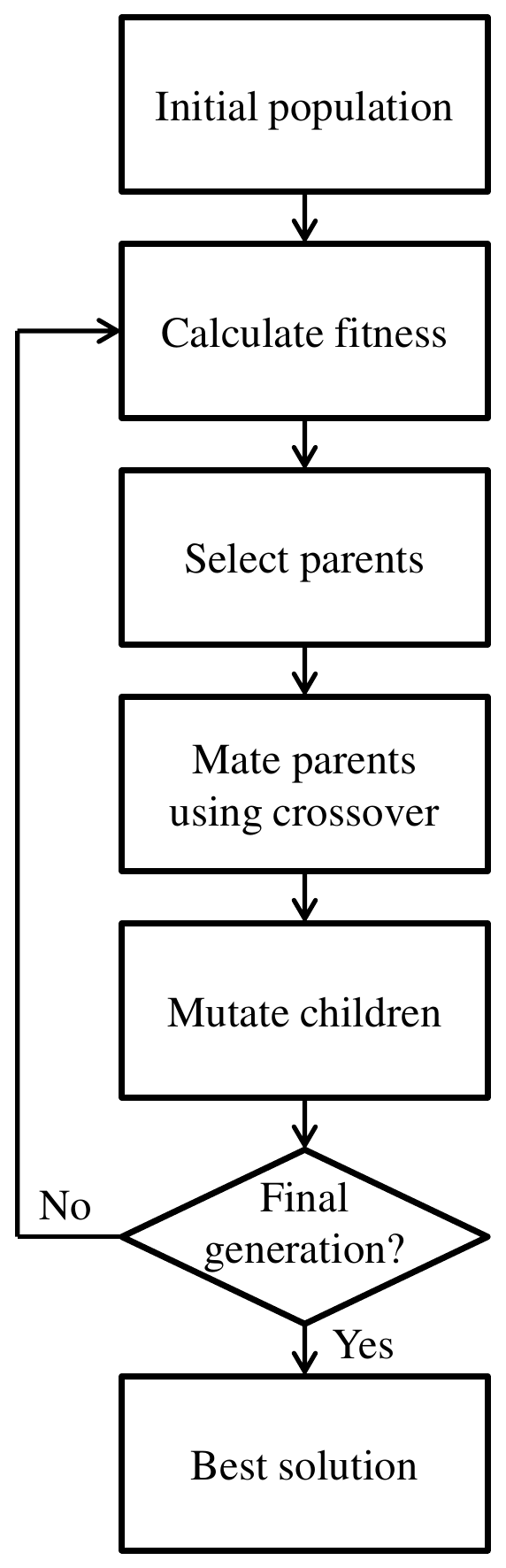}
	
	\caption{Flowchart outlining the steps in the genetic algorithm.  }
	\label{f:genetic}
\end{figure}

Genetic algorithms are a class of meta-heuristics that aim to emulate the process of natural selection. They use a population of chromosomes to represent possible solutions to a problem, with each chromosome consisting of a set of genes that describe the solution. To create a new generation of chromosomes, chromosomes from the current generation are randomly combined in proportion to their fitness. In this way, traits from the strongest chromosomes are most likely to be carried forward to the next generation \citep{Mitchell1998}. Genes are also randomly mutated to avoid being trapped in local optima. This process is outlined in Figure \ref{f:genetic}. A common addition to GAs is elitism, in which the best chromosome(s) from the current generation are propagated to the next generation without modification. Elitism guarantees that solution quality will not decrease \citep{Baluja1995}.

The set of chromosomes in a given population will be denoted by $N$. The rest of this section first discusses possible chromosome representations and how a maintenance schedule is generated from a chromosome, before presenting the methods used in each step of the GA. 

\subsection{Chromosome representation and schedule generation} \label{s:chromosome}

Several different chromosome representations were considered. An obvious representation to choose is a direct representation, where the genes consist of binary variables corresponding to the optimisation variables introduced in Section \ref{s:problem}. However, most of the values that this chromosome representation can take result in infeasible schedules due to the constraints on the availability of workers and maintenance bays. Thus, it was desirable to develop a representation that would reliably produce feasible schedules. 

\begin{figure}
	\centering
	\subfloat[Tasks scheduled in order 1,2]{
		\includegraphics[width=0.8\textwidth]{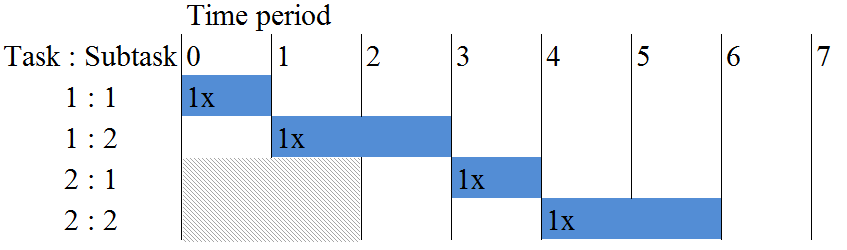}\label{sf:greedy1}
	}
	
	\subfloat[Tasks scheduled in order 2,1]{
		\includegraphics[width=0.8\textwidth]{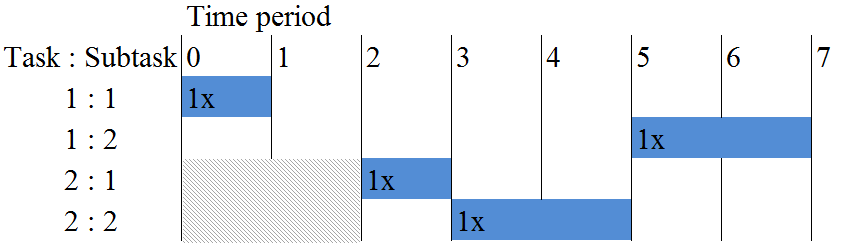}\label{sf:greedy2}
	}
	\caption{This figure shows two tasks being scheduled by the greedy heuristic. Task 1 becomes available for maintenance in time period 0, and task 2 in time period 2 (denoted by the striped area). The tasks are otherwise identical, consisting of two subtasks that each require 1 worker to complete. The first subtask of each task has a duration of 1 time period, and the second subtask has a duration of 2 time periods. For this example, there is only 1 worker available and the tasks do not require maintenance bays. In (a), the tasks are greedily scheduled in the order 1,2, while in (b) the tasks are greedily scheduled in the order 2,1. }
	\label{f:greedy_example}
\end{figure}

The chromosome representation proposed for this application is an ordered list of tasks. The tasks are then scheduled using a greedy heuristic in the order specified by the chromosome to determine the values of the optimisation variables outlined in Section \ref{s:variables}. The greedy heuristic places the subtasks of each task as early as possible while satisfying the constraints in Section \ref{s:constraints}. Provided there is always at least 1 maintenance bay and the minimum number of workers required for any subtasks available in each time period, then this representation is guaranteed to always produce a feasible solution. Figure \ref{f:greedy_example} shows the schedules resulting from using the greedy heuristic on the two possible chromosomes for a scenario with two tasks. 

The greedy heuristic can fail to produce a feasible schedule in certain cases. For example, if the type of worker required for a specific subtask is only available for a small number of time periods at the beginning of the schedule, and the task requires a maintenance bay, then if the task is too late in the chromosome the greedy heuristic can fail to find a valid spot for the subtasks to be placed. 

\subsection{Initial population}

The characteristics of the initial population of chromosomes can have a large impact on the performance of the algorithm \citep{Diaz-Gomez2007}. There is a trade-off between having an initial population that contains good initial solutions, which improves the probability of finding a good final solution \citep{Burke2004}, and having a diverse initial population, which helps avoid premature convergence \citep{Leung1997}. To strike a balance between diversity and quality, two of the chromosomes in the initial population are selected using heuristics, while the remaining are randomly generated. 

The first of the heuristics creates a chromosome by sorting the tasks by their ready times. The reasoning behind this heuristic is that it should avoid situations like Figure \ref{sf:greedy2} where the task with the later ready time is performed in between the subtasks of the other task, leading to a large makespan for the first task. The second heuristic is primarily aimed at producing a feasible schedule in scenarios where many of the chromosomes are unable to be converted into a feasible schedule by the greedy heuristic. Tasks that have subtasks that can only be performed in a limited range of times are placed earlier in the chromosome so that they are likely to be able to be scheduled by the greedy heuristic. 

\subsection{Calculating the fitness of a chromosome}

Two fitness functions were investigated. The first fitness function calculates the fitness of the $n$-th chromosome, $\phi_{n}^{1}$, as:

\begin{equation} \label{eq:fitness1}
\phi^{1}_{n} = \max \{J_{\nu} \; \forall \nu \in N\} - J_{n}
\end{equation}
where $J_{n}$ is calculated using the objective function from Eq.~(\ref{eq:objective}). Note that the max term considers only the chromosomes that result in a feasible schedule. 

The second function calculates the fitness, $\phi^{2}_{n}$, as:

\begin{equation} \label{eq:fitness2}
\phi^{2}_{n} = \frac{1}{J_{n} -  \sum\limits_{i \in I} f_{i}\textrm{min-makespan}_{i}}
\end{equation}

In this formula, the minimum possible makespan cost is subtracted from the value of the objective function. A perfect schedule in which all tasks take the minimum possible time and are completed before their deadlines would therefore result in a denominator value of 0, and a corresponding infinite fitness value. Note that if a perfect schedule is found, this can simply be returned as the best schedule without running the GA to completion. 

\subsection{Parent selection}

A pair of parents are randomly selected for each chromosome in the next generation in proportion to their fitness using the roulette wheel selection method \citep{Davis1991}. Chromosomes that do not result in feasible schedules do not have a fitness and are not considered as valid parents. 

\subsection{Crossover}

\begin{figure}
	\centering
	\subfloat[Before crossover]{
		\includegraphics[width=0.8\textwidth]{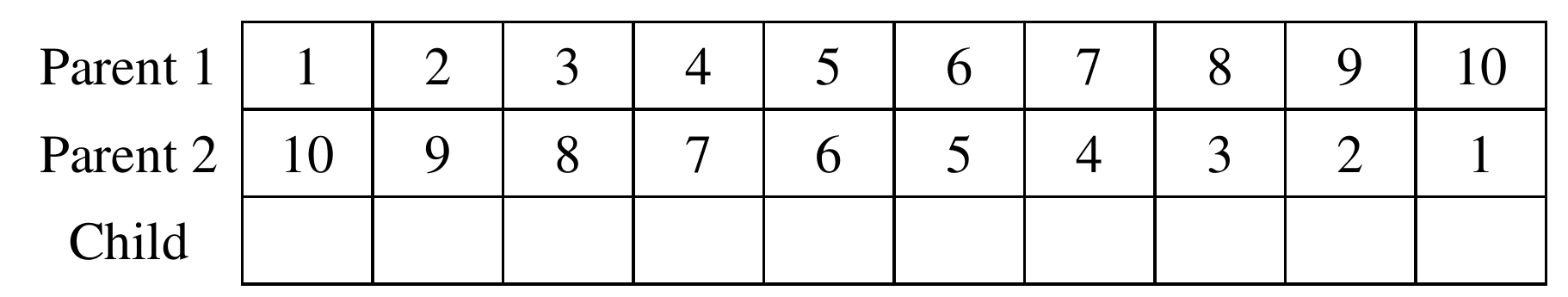}\label{sf:crossover1}
	}
	
	\subfloat[Dominant parent's genes copied across]{
		\includegraphics[width=0.8\textwidth]{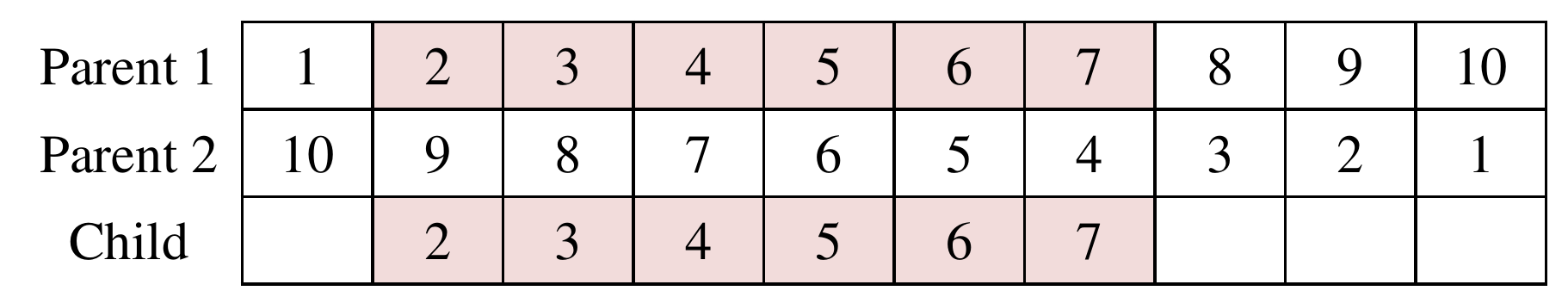}\label{sf:crossover2}
	}
	
	\subfloat[Non-dominant parent's genes copied across]{
		\includegraphics[width=0.8\textwidth]{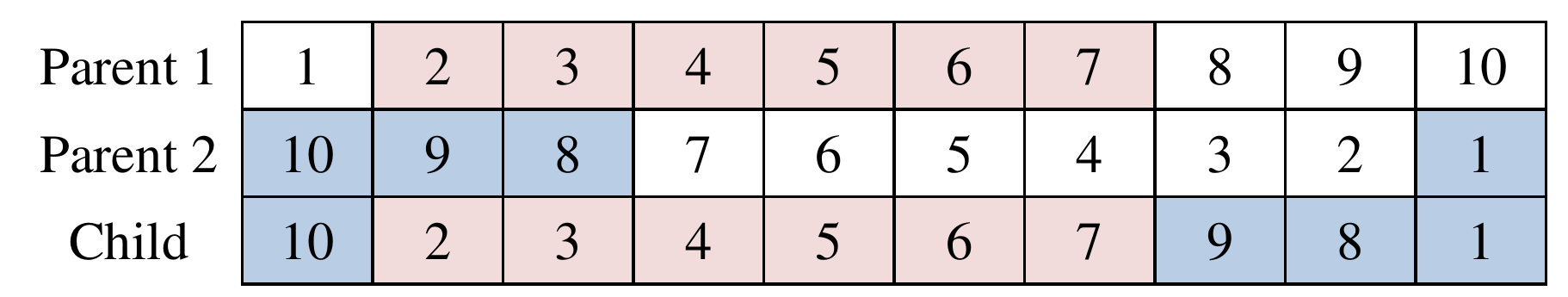}\label{sf:crossover3}
	}
	
	\subfloat[Mutation]{
		\includegraphics[width=0.8\textwidth]{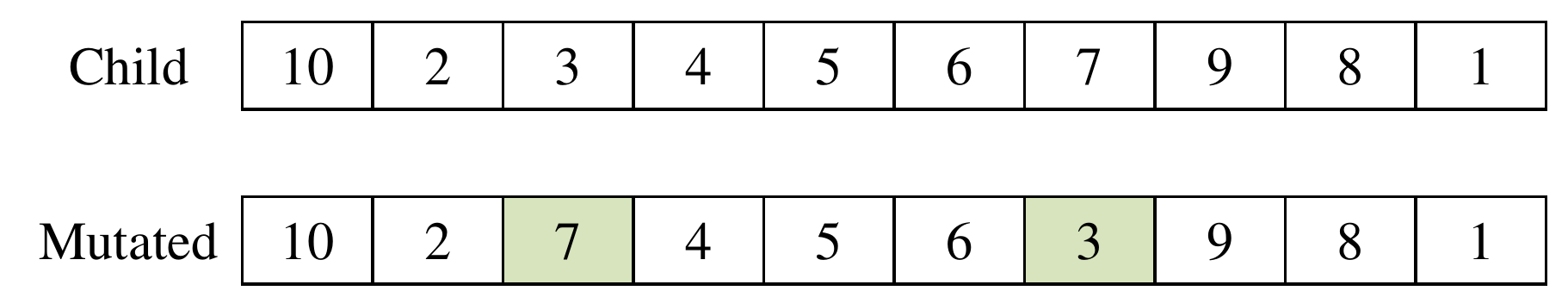}\label{sf:mutation}
	}
	\caption{Demonstration of the crossover and mutation process on chromosomes consisting of 10 genes. Parent 1 is selected as the dominant parent, and crossover points between the 1st and 2nd genes, and between the 7th and 8th genes are chosen. In (b), the genes between the crossover points in parent 1 are copied to the child. In (c), the genes not currently in the child chromosome are inserted in the order they appear in parent 2. Finally, in (d), a random mutation swaps the 3rd and 7th genes. }
	\label{f:crossover}
\end{figure}

Two-point crossover is used to generate a new chromosome from the two parents, as illustrated in Figure \ref{f:crossover}. First, one of the parents is selected at random to be the dominant parent and two crossover points in the chromosome are randomly selected. If the number of tasks between the crossover points is larger than the number of tasks outside the crossover points, then the dominant parent's tasks between the crossover points are copied to the child chromosome. Otherwise, the dominant parent's tasks outside of the crossover points are copied to the child chromosome. The tasks that were not copied across are then greedily placed into the child chromosome in the order in which they appear in the non-dominant parent. 

\subsection{Mutation}

The final step in each round of the GA is to mutate the children. A probabilistic check against the mutation rate is performed for each gene in the chromosome. If the check succeeds, then the gene is swapped with another randomly selected gene in the chromosome. The result of this is shown in Figure \ref{sf:mutation}.  

\section{Computational Study} \label{s:comp_study}

Two weekly maintenance schedules from a sample mine were used to generate the input datasets used in this paper. Each weekly schedule had just under 100 pieces of equipment, and each piece of equipment had between 1 and 50 subtasks, yielding a total of approximately 800 subtasks. 25 different types of people were required for performing the work, each with differing availability levels. Some types of people were only available during the day-shift, while others were available during both the day-shift and night-shift, but in varying numbers. Approximately $1/3^{\textrm{rd}}$ of the pieces of equipment required a maintenance bay, while the remaining were serviced in the field. The mine site under consideration had 5 bays in the maintenance shed available for equipment being serviced. 

This section first evaluates the performance of the commercial solver and GA approaches as the number of tasks is varied, followed by a comprehensive comparison of the methods on randomly generated datasets with the task and worker tightnesses varied. The commercial solver used was Gurobi \citep{Gurobi}, and, unless otherwise specified, a time limit of 600s was used for solving the model. The methods tested were the GA with the fitness function defined in (\ref{eq:fitness1}) (GA1), the GA with the fitness function defined in (\ref{eq:fitness2}) (GA2), the Gurobi (MILP), and Gurobi seeded with an initial solution using a heuristic (MILP+H). The heuristic used in the MILP+H method simply sorted the tasks by their ready time. 

The GA used a population size of 100 chromosomes, 60 generations, a mutation rate of 0.1\% per gene, and elitism of 1 chromosome. These values were experimentally found to give good performance. These parameters were not varied for the different fitness functions as the intention was to show the difference due only to the choice of fitness function. The purpose of this paper was also not to determine the optimal parameters for the GA, as these should be tuned for the specific scenario under consideration. Unless otherwise noted, all calculation times are from an i7-4810MQ with 16GB of RAM, and Gurobi was set to use a maximum of 4 threads. The GA was programmed by the authors in Python and is single threaded. 

\subsection{Varying the number of tasks} \label{s:varying_num_tasks}

\begin{figure}
	\centering
	
	\includegraphics[width=0.8\textwidth]{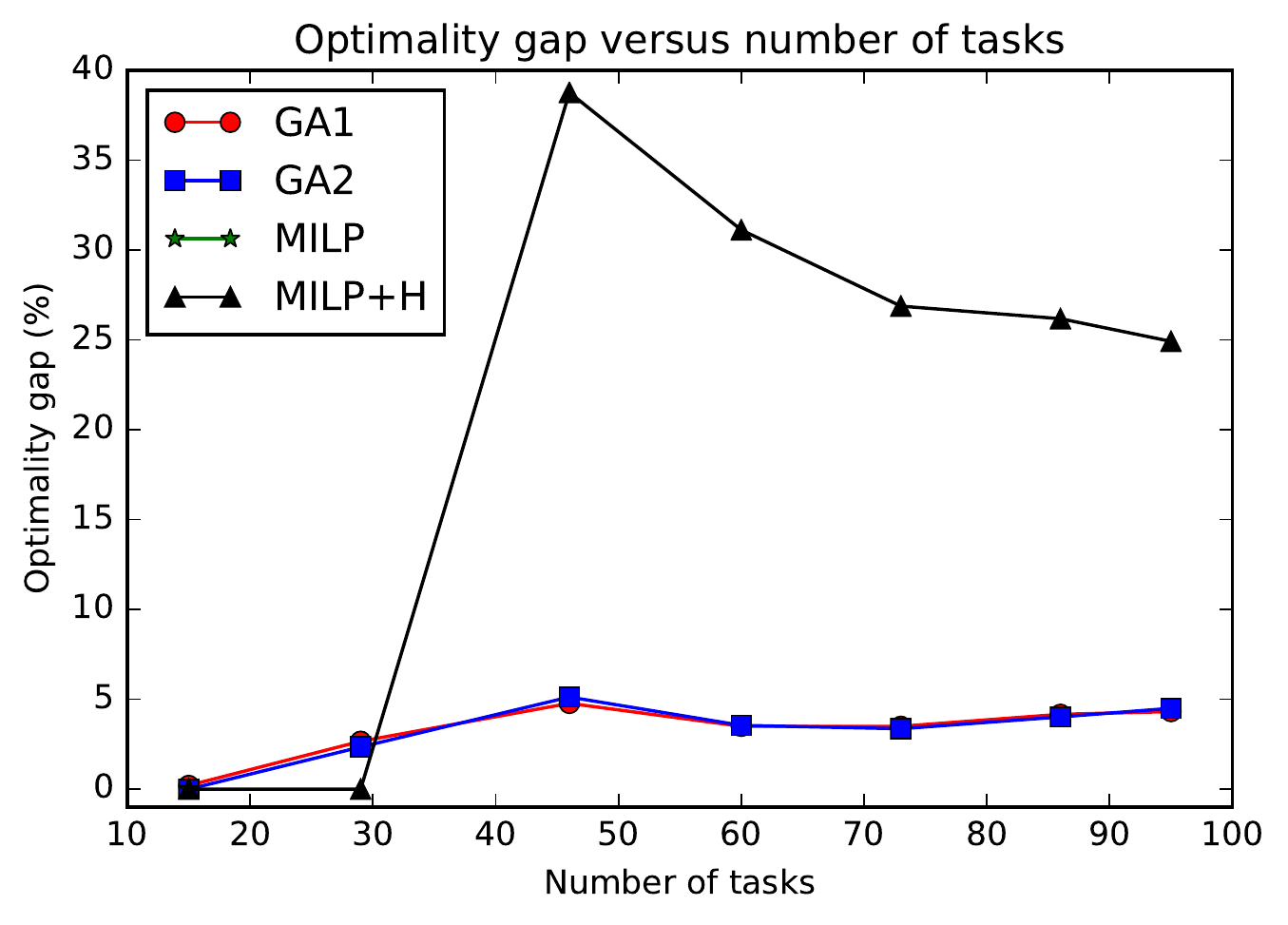}
	
	\caption{Optimality gap versus number of tasks. Note that a feasible solution was only found in the first case for Gurobi solving the MILP. }
	\label{f:varying_num_tasks_optimality_gap}
\end{figure}

The performance of the approaches as the size of the problem was varied was first examined. The number of tasks was varied to correspond to approximately 1, 2, 3, 4, 5, 6, and 7 days worth of tasks in the original schedule. Figure \ref{f:varying_num_tasks_optimality_gap} shows the optimality gap for each method, where the optimality gap was calculated based on the best lower bound calculated by Gurobi using the following formula:

\begin{equation} \label{eq:optimality}
\textrm{optimality gap} = \frac{\textrm{best solution cost}}{\textrm{best bound}} - 1
\end{equation}

\begin{figure}
	\centering
	
	\includegraphics[width=0.8\textwidth]{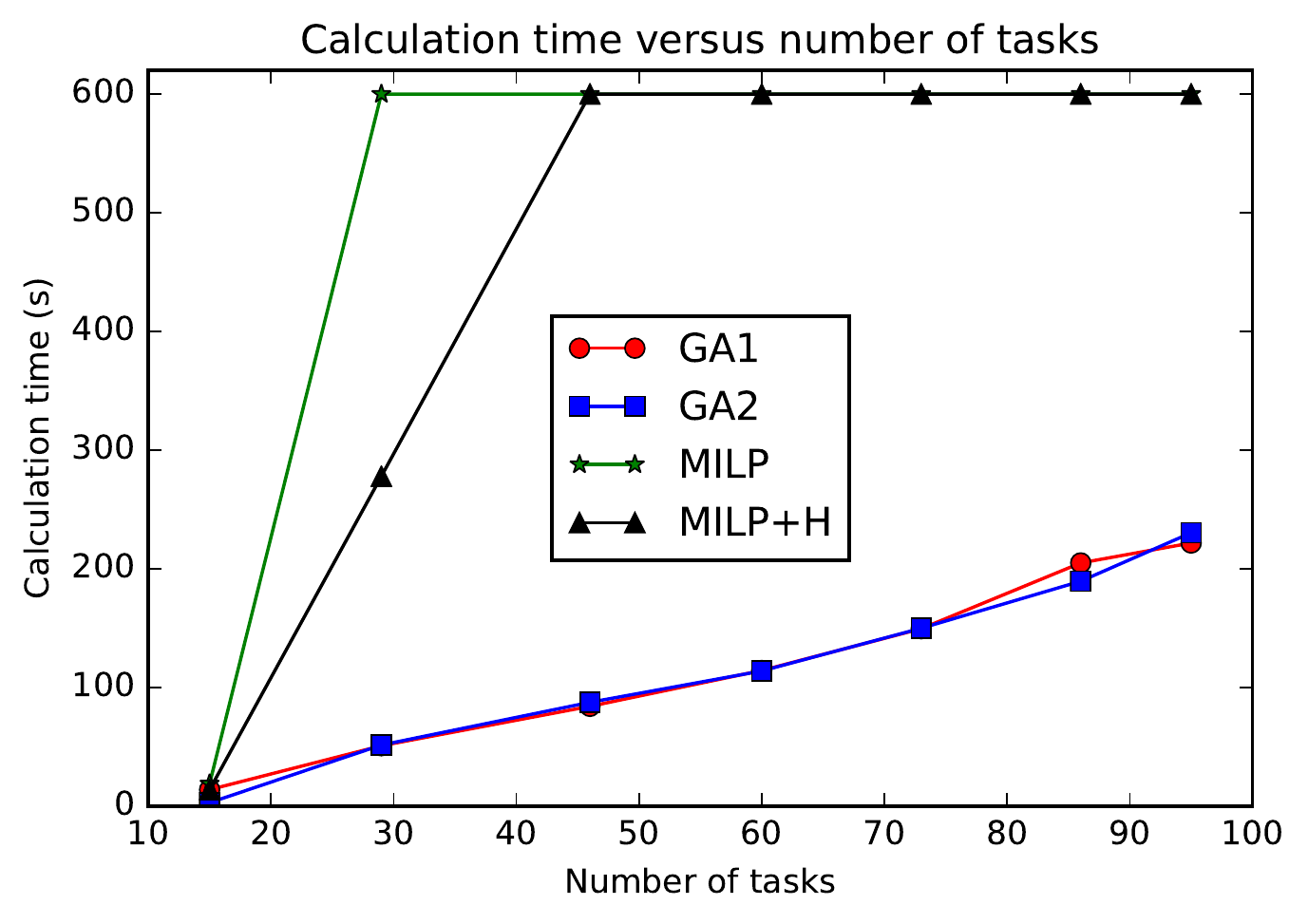}
	
	\caption{Calculation time versus number of tasks. }
	\label{f:varying_num_tasks_calc_time}
\end{figure}

Gurobi was only able to find a solution in the first case, in which it found the optimal solution. When provided with an initial solution, it found the optimal solution in the first two cases, but was unable to improve upon the heuristically generated initial solution in the remaining cases. Figure \ref{f:varying_num_tasks_calc_time} shows the calculation times of each method. As can be seen, Gurobi very quickly hit the time limit of 600s, while the maximum calculation time for the GA approaches was approximately 230s. 

To highlight the infeasibility of the commercial solvers for realistic problem sizes, Gurobi was run for 6 hours using 8 threads on the full 7 days worth of tasks. In this case, the models have over 300,000 binary variables to solve for. Even running for 6 hours, Gurobi was unable to find a feasible solution for the MILP model, or improve from the initial solution for the MILP+H model. In addition, Gurobi was unable to improve the lower bound from the lower bound produced by the initial root relaxation of the model. 

\subsection{Random scenarios}

A set of random scenarios was generated from the input schedules to comprehensively test the performance of each method. Two aspects of the problem were varied---the tightness of the deadline, and the tightness of the worker availability constraints. While the tightness of the deadline impacts the objective value of the optimal schedule, it does not impact on the difficulty of the problem. The tightness of the worker availability constraints, on the other hand, impacts both the objective value of the optimal schedule as well as the problem difficulty. 100 random scenarios were created by randomly sampling the ready time of each task. The deadline of each task was generated from the ready time, minimum possible makespan of the task, and the specified deadline tightness, $\pi$, using the following formula:

\begin{equation}
c_{i} = b_{i} + \pi \times \textrm{min-makespan}_{i}
\end{equation}

Values of $\pi \in \{1,1.5,2\}$ were tested. Three levels of worker tightness were also considered. The tightest level used the actual number of workers available for the supplied schedules---in this case, there were between 1 and 6 workers of most types available, depending on whether it was the day shift of night shift, while 1 or 2 of the types of workers had 10 workers available. The next tightest level assumed that there were 10 workers of every type available. In this way, most of the worker availability constraints become somewhat irrelevant, reducing the difficulty of the problem. Finally, the loosest level of worker tightness assumed that 15 workers of every type available. 

Each of the 100 random scenarios was tested using every combination of deadline tightness and worker tightness, yielding 900 different scenarios in total. A summary of the results are displayed in Tables \ref{t:results} and \ref{t:results2}. Note that the results for the MILP method are omitted as Gurobi was only able to find a feasible solution in 6 of the 900 instances. The heuristic used to seed the MILP+H method produced feasible solutions in all but 3 of the instances, and in 14 cases Gurobi found the optimal solution. It should be noted, though, that the GA1 and GA2 methods also found the optimal solution in those 14 cases, suggesting that they were particularly easy instances. 

Table \ref{t:results} shows the average optimality gap of each method calculated using (\ref{eq:optimality}) and the best lower bound found by Gurobi. It is clear that Gurobi is significantly outperformed by the GA approaches, while, in general, the GA1 approach outperformed the GA2 approach. Similar to the scenarios in Section \ref{s:varying_num_tasks}, Gurobi struggled to improve upon both the heuristically generated initial solution and the initial lower bound. It is therefore difficult to comment on how close to the true optimal solution the GA approaches were. 

\begin{figure}
	\centering
	\subfloat[High cost]{
		\includegraphics[width=0.8\textwidth]{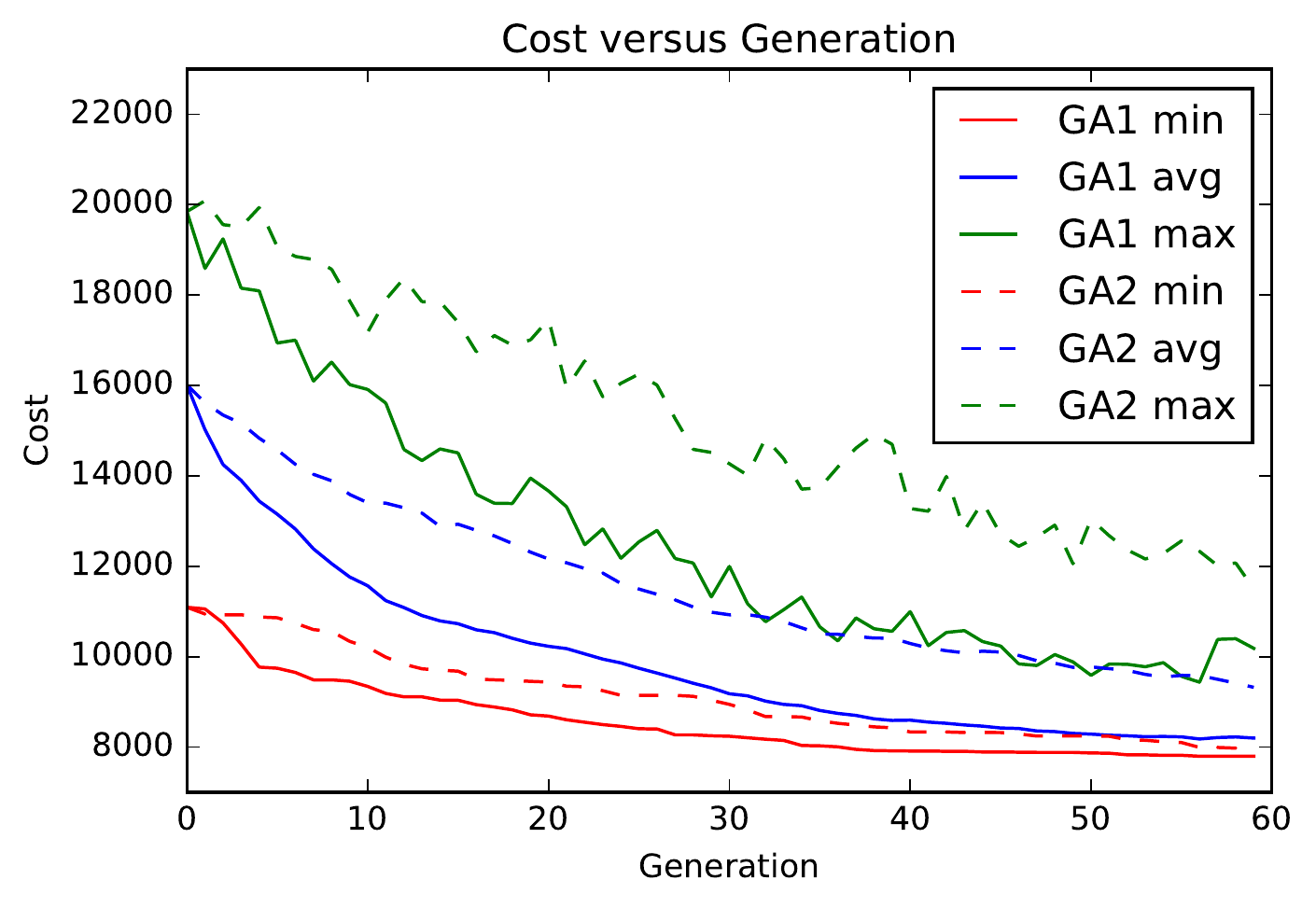}\label{sf:ga_high}
	}
	
	\subfloat[Low cost]{
		\includegraphics[width=0.8\textwidth]{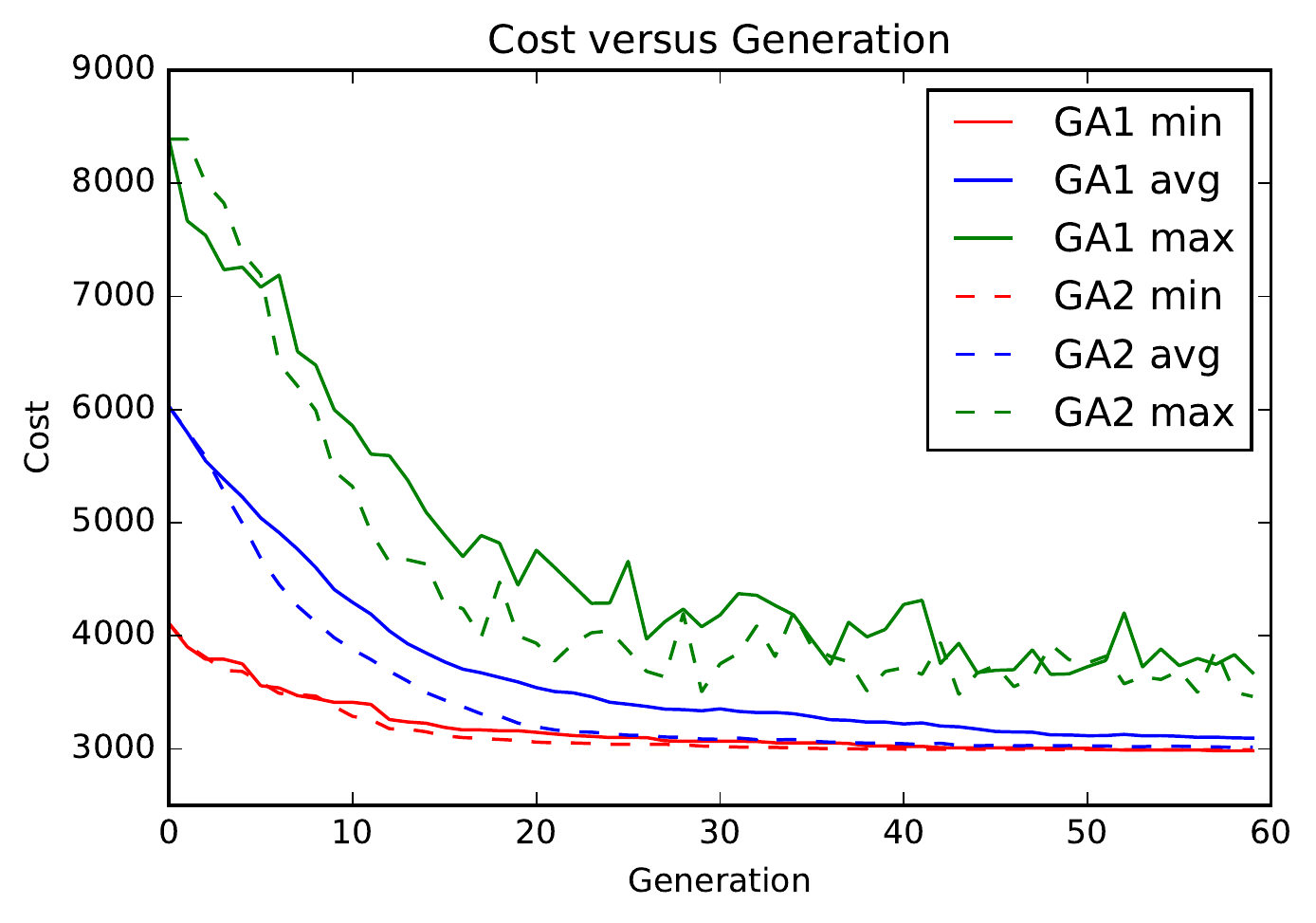}\label{sf:ga_low}
	}
	
	\caption{Performance of the GA1 and GA2 methods as the number of generations is increased. The minimum, average, and maximum costs in each generation are reported. In (a), the solution costs are significantly higher than the lowest possible schedule cost of approximately 2900, while in (b) the solution costs are very close to it.  }
	\label{f:ga_generations}
\end{figure}

Table \ref{t:results2} shows the relative performance of each method using the GA1 method as the reference point. As can be seen, the GA1 method clearly outperforms the GA2 approach when both the deadlines and worker constraints are tight, with an almost 5\% difference in their costs. As these constraints are loosened, the performance difference between the methods decreases. This is a characteristic of the different fitness functions used---when the cost of the schedules are high, the fitness function in (\ref{eq:fitness1}) provides much better discrimination between chromosomes than the fitness function in (\ref{eq:fitness2}). This is highlighted in Figure \ref{sf:ga_high}, where the GA1 approach is shown to converge significantly quicker than the GA2 approach. On the other hand, when the schedule costs are very low, the GA2 method can converge significantly quicker than the GA1 method, as shown in Figure \ref{sf:ga_low}. While GA2 converges faster however, the cost of the best chromosome progresses similarly with the number of generations for both methods. 

\begin{table}
	\caption{Average optimality gap to best known bound for random scenarios}
	\label{t:results}
	\centering
	\begin{tabular}{ c | c c c c }
		\toprule
		 & Deadline tightness & Tight & $\rightarrow$ & Loose \\
		\midrule
		Worker tightness & Method & & & \\
		\midrule
		& MILP+H & 100.1\% & 100.4\% & 63.0\% \\ 
		Tight & GA1 & \textbf{48.6\%} & \textbf{43.1\%} & \textbf{22.1\%} \\ 
		& GA2 & 55.6\% & 46.6\% & 24.0\% \\ 
		\midrule
		& MILP+H & 52.2\% & 42.7\% & 20.7\% \\ 
		$\downarrow$ & GA1 & \textbf{23.5\%} & 15.3\% & \textbf{3.3\%} \\ 
		& GA2 & 25.1\% & \textbf{15.1\%} & 3.6\% \\ 
		\midrule
		& MILP+H & 15.3\% & 16.8\% & 7.5\% \\ 
		Loose & GA1 & \textbf{5.3\%} & \textbf{4.9\%} & \textbf{0.3\%} \\ 
		& GA2 & 5.9\% & \textbf{4.9\%} & 0.6\% \\ 
		\bottomrule
	\end{tabular}
\end{table}

\begin{table}
	\caption{Average performance deficit relative to GA1 for random scenarios. Bold entries are the cases where the GA1 method was outperformed by the respective method. }
	\label{t:results2}
	\centering
	\begin{tabular}{ c | c c c c }
		\toprule
		& Deadline tightness & Tight & $\rightarrow$ & Loose \\
		\midrule
		Worker tightness & Method & & & \\
		\midrule
		& MILP+H & 34.4\% & 39.2\% & 32.6\% \\
		Tight & GA1 & 0.0\% & 0.0\% & 0.0\% \\ 
		& GA2 & 4.9\% & 2.5\% & 1.5\% \\ 
		\midrule
		& MILP+H & 21.6\% & 22.3\% & 16.4\% \\ 
		$\downarrow$ & GA1 & 0.0\% & 0.0\% & 0.0\% \\ 
		& GA2 & 1.3\% & \textbf{-0.1\%} & 0.4\% \\ 
		\midrule
		& MILP+H & 9.2\% & 11.0\% & 7.0\% \\
		Loose & GA1 & 0.0\% & 0.0\% & 0.0\% \\  
		& GA2 & 0.5\% & 0.0\% & 0.2\% \\  
		\bottomrule
	\end{tabular}
\end{table}

\section{Conclusion} \label{s:conc}

This paper developed and compared several methods for automatically generating a maintenance schedule for a typical fleet of mining equipment. Current approaches for scheduling maintenance are manual and time intensive. A mixed-integer linear programming model of the problem was formulated. Commercial optimisation software was unable to sufficiently solve the model, even when supplied with a heuristically generated starting schedule, necessitating the development of alternative strategies. To this end, a genetic algorithm approach was developed, which significantly outperformed the commercial optimisation software. Two fitness functions were also compared, with a linear fitness function shown to in general outperform an inverse fitness function. 

There are several avenues for future work. Further improvements to the performance of the genetic algorithm without increasing the computation time can be achieved by switching from Python to a compiled language such as C++. Genetic algorithms are also embarrassingly parallelisable, so further reductions in the computation time can be achieved in this way. On the algorithmic side, running multiple independent genetic algorithms in parallel may improve the robustness of the approach by helping to avoid local minima. Dynamically switching the fitness function used based on the current chromosome costs could also yield potential improvements in the convergence rate of the algorithm. Finally, methods for incorporating the previous schedule as a starting point when replanning could be investigated. 

\section*{Acknowledgements}

This work was supported by the Rio Tinto Centre for Mine Automation and the Australian Centre for Field Robotics, University of Sydney, Australia.

\bibliographystyle{authordate1} 
\bibliography{ref}

\end{document}